\documentclass[journal, onecolumn, 12pt]{IEEEtran}

\usepackage{amsmath,amssymb,amsfonts,bm}
\usepackage{algorithmic,tikz}
\usepackage{graphicx}
\usepackage{textcomp}
\usepackage{xcolor}
\usepackage{subfigure}
\usepackage{url,hyperref}
\usepackage{authblk}
\usepackage[a4paper, margin=1in]{geometry}
\usepackage{setspace}
\usepackage{multirow}
\usepackage{makecell}
\usepackage{flushend}
\usepackage{chngcntr}
\usepackage{booktabs,dcolumn}
\usepackage{enumitem}
\newcolumntype{d}[1]{D{.}{.}{#1}} 
\renewcommand{\ast}{{}^{\textstyle *}} 

\author{Lhuqita Fazry}
\author{Valentino Vito}
\affil{Faculty of Computer Science, Universitas Indonesia, Depok, Indonesia\\
lhuqita.fazry@ui.ac.id \qquad valentino.vito11@ui.ac.id}

\begin{document}

\title{Unsupervised Raindrop Removal from a Single Image using Conditional Diffusion Models}

\maketitle

\begin{abstract}
Raindrop removal is a challenging task in image processing. Removing raindrops while relying solely on a single image further increases the difficulty of the task. Common approaches include the detection of raindrop regions in the image, followed by performing a background restoration process conditioned on those regions. While various methods can be applied for the detection step, the most common architecture used for background restoration is the Generative Adversarial Network (GAN). Recent advances in the use of diffusion models have led to state-of-the-art image inpainting techniques. In this paper, we introduce a novel technique for raindrop removal from a single image using diffusion-based image inpainting.
\end{abstract}


\section{Introduction}
Rain is a weather condition that can adversely affect the quality of captured images. Due to rain, droplets and streaks of water can obscure the aforementioned images. Deraining is the act of removing these rain elements from an image or video, usually by machine and deep learning methods. Two types of rain elements are potentially present: (1) rain streaks, which are line-shaped in appearance \cite{quan2019}, and (2) raindrops. In this paper, we focus on the task of raindrop removal from a single image. Hence, video restoration is beyond the scope of this study, and rain streaks are assumed to not be present in the images.

\begin{figure}[t]
    \centering
    \includegraphics[width=\textwidth]{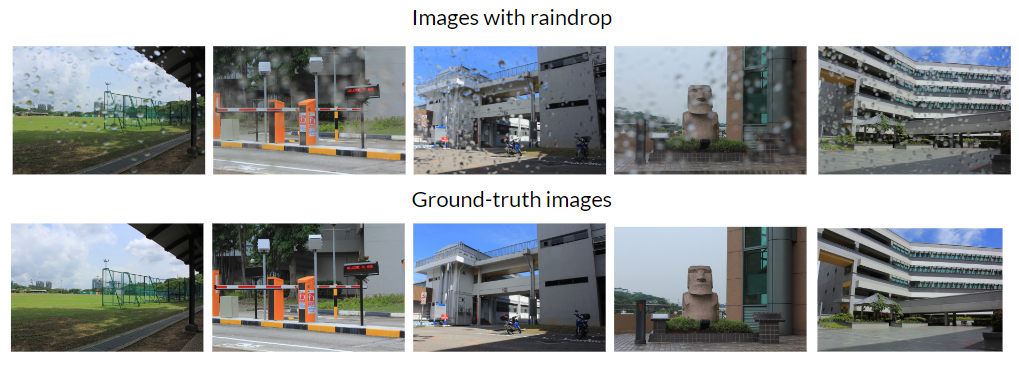}
    \caption{Images with raindrops along with their corresponding ground-truth clean images from the Raindrop dataset \cite{qian2018}}
    \label{fig:raindrop_images}
\end{figure}

Raindrops occurring in an image are formed by rays of reflected light from a wider environment \cite{qian2018}. The circular regions produced by raindrops give a blurry effect to the image, which degrades the overall image quality. Raindrop removal on a single image is an image restoration task that was introduced by Qian et al. \cite{qian2018}. They released a novel raindrop removal dataset, known as the \textit{Raindrop dataset}, for experimentation purposes (see Figure \ref{fig:raindrop_images}). Afterward, they applied an attentive Generative Adversarial Network (GAN) to this dataset. We utilize a diffusion model in this study instead of a GAN.

Diffusion models are generative models capable of producing high-quality images. They are latent variable models borrowing concepts from nonequilibrium
thermodynamics \cite{sohl-dickstein_deep_2015}. Diffusion models have gained much traction in recent years as they have been shown to outperform GANs in image synthesis \cite{dhariwal2021}. Notably, they were recently applied to image inpainting \cite{lugmayr_repaint_2022}, an image restoration task that aims to artificially fill in missing sections within an image. Inpainting is closely related to raindrop removal, especially in regard to filling in the blurry circular regions left by the raindrops in order to reconstruct the background image.

To address the problem of raindrop removal from a single image, we propose \textbf{DropWiper}. DropWiper is a novel two-step architecture consisting of raindrop detection and background reconstruction. Although raindrop detection merely uses a convolutional neural network as the raindrop mask generator, the background reconstruction uses the aforementioned diffusion model that was used for inpainting. Two datasets are used for training and testing our model: the Raindrop dataset \cite{qian2018} and the Cityscapes dataset \cite{Cordts2016Cityscapes}. For training purposes, we also add synthetic raindrops onto the Cityscapes images via a non-parametric refraction model.

\section{Related Work}
\subsection{Raindrop Removal}
Raindrop removal has seen several applications, such as in unmanned aerial vehicle inspection \cite{xu2021} and coastal video enhancement \cite{kim2020}. Most approaches toward raindrop removal rely on GANs \cite{xia_raindrop_2022,nguyen2021,uzun2019,yang2021,shao2021,xu2021}. Most recently, Xia et al. \cite{xia_raindrop_2022} devised a two-step GAN to maintain balance between raindrop removal and image inpainting. Their strategy is quite similar to our own but differs in the use of architecture. In contrast to GANs, we employ diffusion models to perform the inpainting. Recently, {\"O}zdenizci and Legenstein \cite{ozdenizci_restoring_2022} designed TransWeather, a patch-based diffusion model for vision restoration in adverse weather conditions. Unlike this study, they approach multiple weather conditions at once, including rain, snow, and haze. Although we use a similar diffusion approach, we are strictly focused on raindrop-filled images.

\subsection{Diffusion Models}
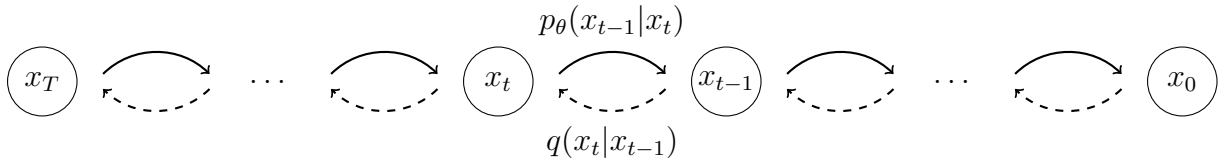
\begin{figure}[t]
    \centering
    \begin{tikzpicture}[x=10mm,y=10mm]
        \draw[fill=white] (0,0) circle (13pt);
        \draw[fill=white] (6,0) circle (13pt);
        \draw[fill=white] (-3,0) circle (13pt);
        \draw[fill=white] (-9,0) circle (13pt);

        \draw[->,thick,black] (-8.2,0.1) to [out=45,in=135] (-6.8,0.1);
        \draw[->,thick,black,dashed] (-6.8,-0.1) to [out=-135,in=-45] (-8.2,-0.1);
        \draw[->,thick,black] (-5.2,0.1) to [out=45,in=135] (-3.8,0.1);
        \draw[->,thick,black,dashed] (-3.8,-0.1) to [out=-135,in=-45] (-5.2,-0.1);
        \draw[->,thick,black] (-2.2,0.1) to [out=45,in=135] (-0.8,0.1);
        \draw[->,thick,black,dashed] (-0.8,-0.1) to [out=-135,in=-45] (-2.2,-0.1);
        \draw[->,thick,black] (0.8,0.1) to [out=45,in=135] (2.2,0.1);
        \draw[->,thick,black,dashed] (2.2,-0.1) to [out=-135,in=-45] (0.8,-0.1);
        \draw[->,thick,black] (3.8,0.1) to [out=45,in=135] (5.2,0.1);
        \draw[->,thick,black,dashed] (5.2,-0.1) to [out=-135,in=-45] (3.8,-0.1);

        \node at (-1.5,0.8) {$p_{\theta}(x_{t-1}|x_t)$};
        \node at (-1.5,-0.8) {$q(x_t|x_{t-1})$};
        \node at (-6,0) {$\dots$};
        \node at (3,0) {$\dots$};
        \node at (-9,0) {$x_T$};
        \node at (0,0) {$x_{t-1}$};
        \node at (-3,0) {$x_t$};
        \node at (6,0) {$x_0$};
        
    \end{tikzpicture}
    
    \caption{The diffusion model. The solid arrows denote reverse processes, whereas the dashed arrows denote forward processes}
    \label{fig:diffusion}
\end{figure}

The theory of diffusion models is developed in detail by Ho et al. \cite{ho2020}. A diffusion model is essentially a Markov chain containing $T$ latent variables $x_1, \dots, x_T$ with respect to the original image data $x_0$. The model consists of (1) a \textit{forward process}, which gradually adds noise to $x_0$ and returns the latent variables in succession, and (2) a \textit{reverse process} for denoising, which gradually removes noise from the latent variables to eventually reobtain $x_0$. The forward process is represented as distributions $q(x_t)$, and the reverse process is represented as distributions $p_\theta(x_t)$. The illustrating diagram is provided in Figure \ref{fig:diffusion}.

For the forward process, the conditional probability of $x_t$ given $x_{t-1}$ can be calculated using the following formula:
\begin{align}
      q(x_t | x_{t-1}) = \mathcal{N}(x_t; \sqrt{1-\beta_t}x_{t-1}, \beta_t \mathbf{I})\label{eq:single_diffusion}
\end{align}
where $\beta_t$ is a variance hyperparameter for the forward process. Let $\alpha_t = 1 - \beta_t$ and $\overline{\alpha}_t = \prod_{s=1}^t \alpha_s$. The conditional probability of $x_t$ given $x_0$ can be directly calculated using the following formula:
\begin{align}
     q(x_t|x_0) = \mathcal{N}(x_t; \sqrt{\overline{\alpha}_t}x_0, (1-\overline{\alpha}_t)\mathbf{I}) \label{eq:direct_diffusion}
\end{align}
For the reverse process, the conditional probability of $x_{t-1}$ given $x_t$ can be calculated using the following formula:
\begin{align}
     p_{\theta}(x_{t-1}|x_t) = \mathcal{N}(x_{t-1}; \mu_{\theta}(x_t, t), \Sigma_{\theta}(x_t, t)) \label{eq:reverse_model}
\end{align}

Diffusion models have numerous applications to image synthesis \cite{rombach2022,saharia2022,ramesh2022}, in which they have become the state-of-the-art approach. In addition, they have been applied to anomaly detection \cite{wyatt2022} and image restoration \cite{ozdenizci_restoring_2022}. This indicates the versatility and effectiveness of diffusion models in the field of image processing.
 
\section{Methodology}
This paper proposes a novel method to remove raindrops from a single image. We call our method DropWiper\footnote{Source code: \url{https://github.com/lhfazry/DropWiper}}. For testing purposes, we use the Raindrop dataset from Qian et al. \cite{qian2018}. This dataset consists of pairs of images. Each pair consists of an image with raindrops and its corresponding ground-truth clean image. Unfortunately, the dataset does not contain ground-truth raindrop masks to identify areas containing raindrops in the image. These masks play an important role to ensure successful raindrop removal. To handle this issue, we divide our method into two steps: pseudo raindrop mask generation and background reconstruction.

In the first step, we create pseudo raindrop masks for the raindrop-filled images in the dataset. Using these pseudo-masks as guidance, we then try to recover the background using the Denoising Diffusion Probabilistic Model (DDPM) \cite{ho2020}, which was recently used for image inpainting \cite{lugmayr_repaint_2022}. This image inpainting method does not expect a \textit{fine-grained} level of masking since it can recover the background even when the masked regions are larger than expected. We design several approaches to generate the pseudo raindrop masks to see which one performs best.

\subsection{Pseudo Raindrop Mask Generation}
A raindrop mask is a binary image that maps areas containing raindrops. This map separates the raindrop areas and the background areas of the image. If an area contains raindrops, then the pixels in the are set to $255$ (white). For the background areas that do not contain raindrops, pixels are set to $0$ (black).

We consider two approaches for generating a pseudo raindrop mask: residual mask generation and synthetic mask generation. In the first approach, we create a \textit{residual mask} simply by subtracting the clean image from the corresponding image with raindrops. Then, we set a threshold to extract the raindrop regions from the resulting residual image. Residual pixels that lie above the threshold are categorized as part of the raindrop area, whereas those that do not are categorized as part of the background. Unfortunately, directly calculating the residual on images often results in poor masking due to the existence of extraneous variables. These variables include brightness, light intensity, other objects, etc. So, we employ several pre-processing techniques to minimize the influence of these variables, including histogram equalization and photometric distortion. In the second approach, we employ a more sophisticated method to generate \textit{synthetic masks}. We then train a supervised raindrop detector by supplying it with a synthetic raindrop ground truth.

\subsubsection{Direct Thresholding on the Residual} \label{subsec:direct_threshold}
Each data point in the Raindrop dataset \cite{qian2018} comes with a pair of images: an image with raindrops and its clean version. We are given two images $A, B \in \mathbb{R}^{H \times W \times 3}$ taken from the dataset, where $A$ is an image with raindrops while $B$ is the corresponding clean image. The notations $H$ and $W$ represent the image's height and weight respectively, while $3$ is the corresponding number of channels for RGB (Red, Green, and Blue) images. We compute the residual image from $A$ and $B$ via one of the following options:
\begin{enumerate}[label=\alph*]
    \item Compute the residual image by subtracting $B$ from $A$. The residual image is defined as follows:
    \begin{align}
        R = A - B
    \end{align}
    \item Compute the residual image by using the absolute difference between $A$ and $B$ as follows:
    \begin{align}
        R = \left\lvert A - B \right\rvert
    \end{align}
    \item Convert $A$ and $B$ into gray-scale images and then define the residual as follows:
    \begin{align}
        R = A - B
    \end{align}
    \item Convert $A$ and $B$ into gray-scale images and then compute the residual image by using the absolute difference between $A$ and $B$ as follows:
    \begin{align}
        R = \left\lvert A - B \right\rvert
    \end{align}
\end{enumerate}

For options $a$ and $b$, we convert the resulting residual into a gray-scale image. Afterward, for all options, we apply a threshold to categorize the pixels. We perform experiments with various threshold values to find the best one. The threshold values are set to $30$, $80$, $120$, and $200$.



\subsubsection{Raindrop Detection}
During this step, we train a model to detect raindrops from an image. Given an image with raindrops, the model outputs a binary mask that segments the raindrops and the background. We use a lightweight Convolutional Neural Network (CNN) as the model. It only has $493$k parameters in total. 

The network has three main parts. The first part consists of two convolution blocks. Each block is a single 2D CNN with batch normalization \cite{Ioffe2015BatchNA} and \texttt{LeakyReLU} \cite{Xu2015EmpiricalEO} as the non-linear activation function. Each CNN uses a kernel of size $3 \times 3$ with stride $2$. The feature dimension of the input image is increased from $3$ to $32$ (by the first block) and then to $64$ (by the second block) to ensure that the model learns richer features. Still in this first part, we add another 2D CNN to further increase the feature dimension to $256$ by a $1 \times 1$ kernel with stride 2, but without the use of batch normalization nor non-linear activation. 

\begin{figure}
    \centering
    \includegraphics[width=0.8\textwidth]{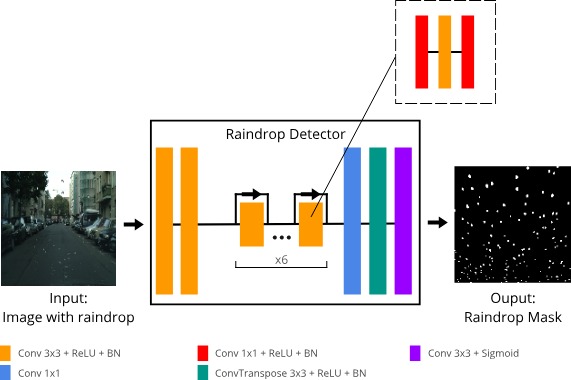}
    \caption{The raindrop detector network consists of $2$ convolutional blocks, followed by $6$ residual blocks, and ends up with $3$ up-sampling blocks. It takes an image with raindrops as input and then returns the raindrop mask as output}
    \label{fig:raindrop_detector}
\end{figure}

For the second part, we employ $6$ residual blocks. Each residual block contains $3$ layers of 2D CNNs equipped with batch normalization and \texttt{LeakyReLU} for non-linear activation. The first CNN takes as input a feature of dimension $256$ and uses a kernel of size $1 \times 1$. The output feature has dimension $64$. In the second CNN, the output dimension continues to be $64$, but we increase the kernel size to $3 \times 3$. The last CNN restores the feature dimension to $256$ using a $1 \times 1$ kernel. In these residual blocks, we force the network to learn spatial patterns since the raindrops are spatially dependent. We stack $6$ blocks and employ a residual connection \cite{he_deep_2015} in each block. A residual connection can be formally defined as $y = \text{model}(x) + x$. The aim of residual connections is to create a skip connection so that forward and backward operations can benefit from it. Residual connections have been proven to stabilize training even in very deep networks.

The final part of the raindrop detector consists of a convolution block, a convolution transpose block, and a single CNN. The convolution block used in this part is the same as the convolution block used in the first part. This block reduces the feature dimension from $256$ to $64$. Then, the convolution transpose block increases the spatial resolution of the image to match the input resolution by up-sampling, and at the same time reduces the feature dimension to $32$. In the very last layer, a CNN is used to further reduce the feature dimension to $1$, denoting a gray-scale feature. The \texttt{sigmoid} function is then applied to obtain a probability value for every pixel. If the value is greater than $0.5$, then the corresponding pixel is categorized as a raindrop. Otherwise, it is categorized as background. Figure \ref{fig:raindrop_detector} illustrates the model architecture.

\begin{figure}
    \centering
    \includegraphics[width=0.8\textwidth]{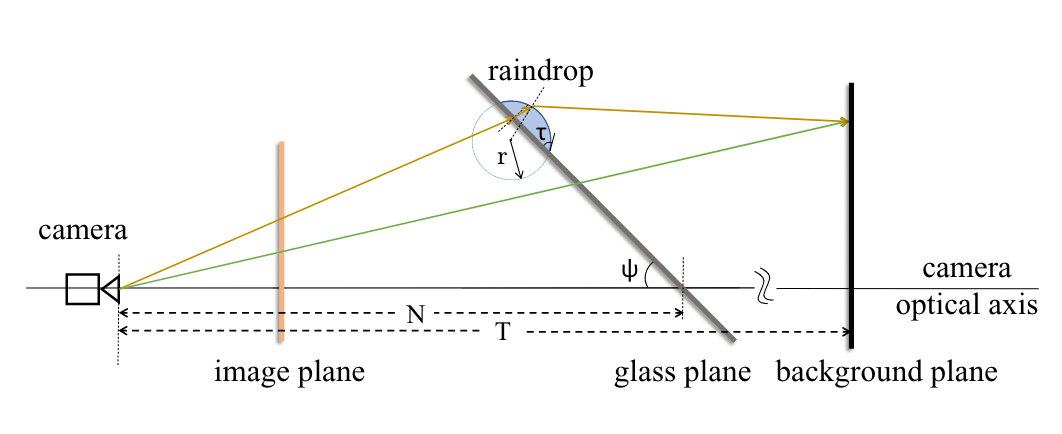}
    \caption{The refraction model. This model is a raindrop modeling technique based on ray tracking. Based on this model, the background image under the raindrop is a refracted manifestation obtained according to some degree and distance}
    \label{fig:refraction_model}
\end{figure}

The big question is how to train this model. The Raindrop dataset from Qian et al. does not contain ground-truth masks. Without raindrop masks, we cannot train the model. We overcome this issue by following the literature \cite{halimeh_raindrop_2009, koch_realistic_2011, bernuth_rendering_2018, you_waterdrop_2016, hao_learning_2019} to create synthetic raindrops using a refraction model. Under this model, the background of the raindrops can be seen as a contracted version of the full background. Figure \ref{fig:refraction_model} illustrates the refraction model. To generate synthetic raindrops, we need several parameters surrounding the camera, including the degree and distance between the camera and the screen. We generate these synthetic raindrops on the Cityscapes dataset \cite{Cordts2016Cityscapes} for two reasons.  First, the Cityscapes dataset provides us with image data and the camera parameters we need. The camera parameters enable us to generate synthetic raindrops using the refraction model. Second, the images contained in the Cityscapes dataset and the Raindrop dataset are quite similar. The Cityscapes dataset contains images of roads, buildings, houses, etc. The Raindrop Dataset has similar venues, including roads, buildings, and campuses.

The refraction model is non-parametric. It is a single forward algorithm without the need for training. The model takes a single clean image as input and then returns two images: an image with synthetic raindrops and the corresponding raindrop mask. Both are then used to train our raindrop detector. The raindrop detector takes an image with synthetic raindrops as input and returns the raindrop mask prediction. The error between the model's prediction and the ground-truth raindrop mask is used to backpropagate the model's weights. After the detector model has been trained, it is applied to detect raindrops on the Raindrop dataset.

\subsection{Background Reconstruction}
Given a pseudo-mask $m$ generated from the previous module, we aim to restore the background on the masked areas. In this module, we employ a pre-trained DDPM model \cite{dhariwal2021} that has been trained on clean images of the Raindrop dataset. Specifically, we run the inference process using the pre-trained model to restore the background. We expect that the model can learn the distribution of clean images so it can generate novel images from that distribution. The intuition is simple. If the model can generate such novel images, then it can be used to reconstruct the background image when conditioned on the raindrop mask. 

Background reconstruction based on a mask essentially consists of predicting the missing pixels due to raindrops conditioned on the mask. The Raindrop dataset comes with ground-truth clean images, so we can evaluate the result using these ground truths. First, we denote $x$ as the ground-truth image, $m \odot x$ as the missing pixels, and $(1-m) \odot x$ as the known pixels. The operator $\odot$ denotes element-wise multiplication.

\begin{figure}
    \centering
    \includegraphics[width=\textwidth]{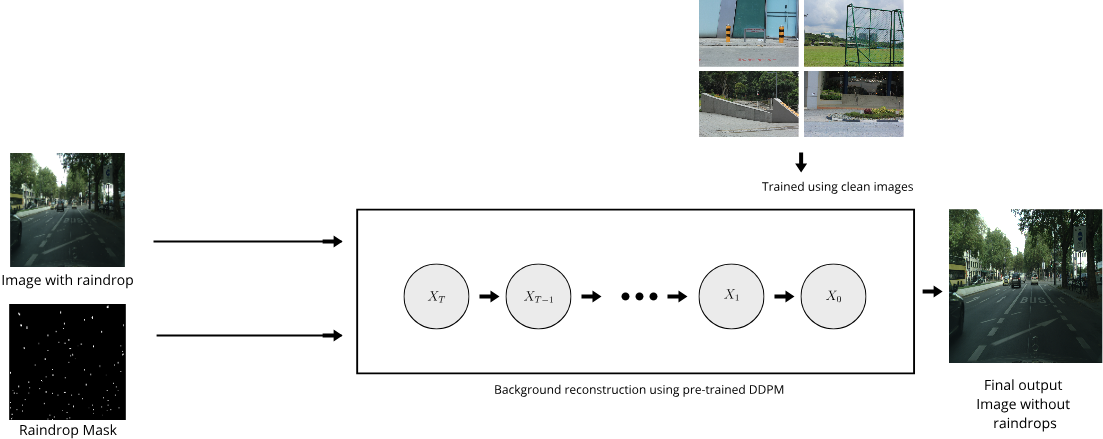}
    \caption{The background restoration process. Given an image with raindrops and a pseudo-mask, the model reconstructs regions identified by the mask}
    \label{fig:background_reconstruction}
\end{figure}

Recall that from the DDPM perspective, model training is a forward diffusion process, while model inference is a reverse diffusion process. From Equation \ref{eq:reverse_model}, we can see that in a single reverse step, we only need the variable $x_t$ to reconstruct $x_{t-1}$. We can ensure that DDPM constructs only on the masked regions via a simple trick. After we obtain $x_{t-1}$, we can alter the known pixels in $(1-m) \odot x_{t-1}$ using the corresponding pixels from $x_{t-1}$ that are obtained by forward diffusion. Fortunately, we can directly go to $x_t$ from $x_0$ for any $t$ by using Equation \ref{eq:direct_diffusion}. So, we conduct the background restoration by combining two directions: forward and reverse. We use Equation \ref{eq:reverse_model} to obtain the unknown pixels and Equation \ref{eq:direct_diffusion} to obtain the known pixels. This technique can be formally defined as follows:
\begin{align}
    x_{t-1} = m \odot x_{t-1}^{\text{known}} + (1-m) \odot x_{t-1}^{\text{unknown}}
\end{align}
where $x_{t-1}^{\text{known}} \sim \mathcal{N}\left( \sqrt{\overline{\alpha}}x_0, (1-\overline{\alpha}_t) \mathbf{I}\right)$ and $x_{t-1}^{\text{unknown}} \sim \mathcal{N}\left(\mu_{\theta}(x_t,t), \Sigma_{\theta}(x_t,t)\right)$. Figure \ref{fig:background_reconstruction} illustrates the process of background restoration.

\section{Experiments}
\subsection{Experimental Datasets}
In the experiments, we use two datasets: the Raindrop dataset by Qian et al. \cite{qian2018} and the Cityscapes dataset \cite{Cordts2016Cityscapes}. The former dataset, Raindrop, contains images with raindrops and their corresponding clean images as ground truth. Our main goal is to detect and remove raindrops from these images and evaluate the results against clean images. This dataset consists of $861$ train data, $58$ validation data, and $249$ test data. Each image has a spatial resolution of $720 \times 480$ with \texttt{RGB} channels. Most images in this dataset are photos taken around campus.

The latter dataset is the Cityscapes dataset. We use the \texttt{gtFine\_trainvaltest.zip} package contained in this dataset. This package contains $5000$ images, along with parameter settings on the camera that is used to take the photos. We employ this dataset to train our model. We generate synthetic raindrops on this dataset and then use the result to train the raindrop detector.



\subsection{Experimental Settings}
We use 1 GPU V100 for all experiments. We implement the raindrops synthetic generator using the \texttt{C++} programming language, adapting the code from Hao et al. \cite{hao_learning_2019}. To better handle the image data using \texttt{C++}, we use \texttt{OpenCV}, an open-source \texttt{C++} library that includes several hundreds of computer vision algorithms.

We implement the raindrop detector using the \texttt{Python 3.8} programming language and the \texttt{PyTorch 1.12} library to better create the model architecture. For easier model training and fine-tuning, we use the \texttt{PyTorch Lightning} framework \cite{Falcon_PyTorch_Lightning_2019}. We train the model for $100$ epochs with batch size $32$. We also use the \texttt{AdamW} optimizer \cite{Loshchilov2017DecoupledWD} with learning rate $10^{-3}$ and weight decay $10^{-4}$. For the learning rate scheduler, we use \texttt{StepLR} with \texttt{step\_size} $5$. We train the model to minimize the \texttt{binary cross entropy} loss as the raindrop detector task is essentially a segmentation task utilizing per-pixel classification.

For the background reconstructor, we use the DDPM model implemented by Dhariwal et al. \cite{dhariwal2021}. We train the model on the clean images of the Raindrop dataset for 1M iterations. We set the number of diffusion steps to $T = 1000$. To save memory, we center-crop the images to size $128 \times 128$. We use the \texttt{AdamW} optimizer with learning rate $10^{-4}$ without weight decay. The training process took 6 days in total.

\subsection{Results and Discussion}
\subsubsection{Residual Masks}
\begin{figure}
    \centering
    \includegraphics[width=\textwidth]{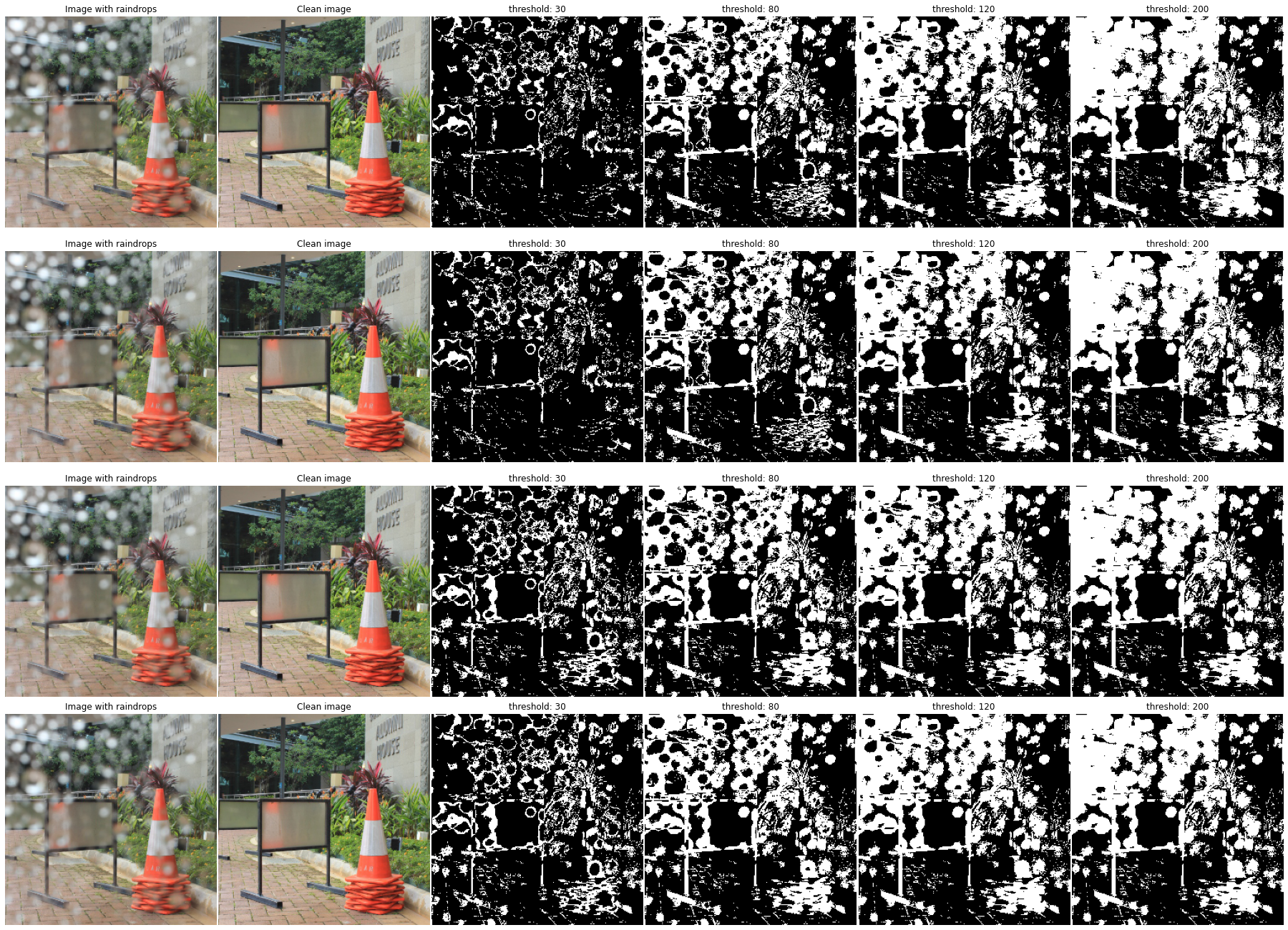}
    \caption{Results of residual masks at different settings. In rows 1 to 4, various residual settings are applied as explained in Subsection \ref{subsec:direct_threshold}. Column 1 and column 2 represent an image with raindrops and its corresponding clean image, respectively. Columns 3 to 6 represent raindrop masks obtained through different threshold settings ($30$, $80$, $120$, and $200$, respectively)}
    \label{fig:residual_masks}
\end{figure}

For this part, we conducted some experiments to generate residual masks based on different settings as explained in Subsection \ref{subsec:direct_threshold}. Figure \ref{fig:residual_masks} shows the results from applying these settings.

It can be seen from Figure \ref{fig:residual_masks} that there is no significant difference between masks obtained from different residual settings. But, we found significant changes after applying different threshold settings since higher threshold values correlate with the increase of masked regions.

\subsubsection{Synthetic Raindrops}
\begin{figure}
    \centering
    \includegraphics[width=\textwidth]{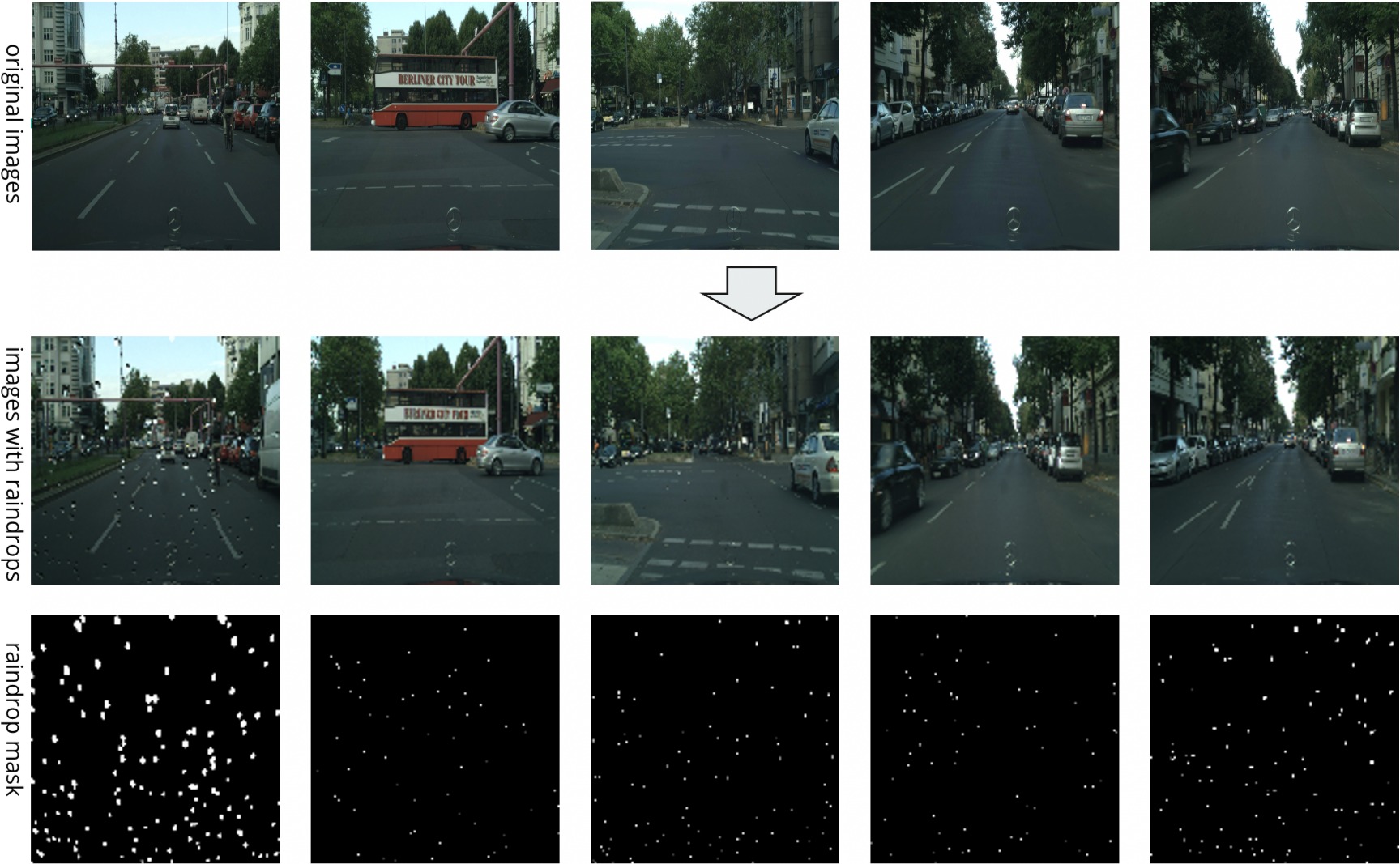}
    \caption{Sample images with synthetic raindrops generated from the refraction model. The refraction model takes a clean image as input and returns an image with raindrops along with its corresponding raindrop map}
    \label{fig:raindrop_synthetic}
\end{figure}

Figure \ref{fig:raindrop_synthetic} shows samples of images obtained through synthetic raindrop generation. 

\subsubsection{Raindrop Detector}
\begin{figure}
    \centering
    \includegraphics[width=\textwidth]{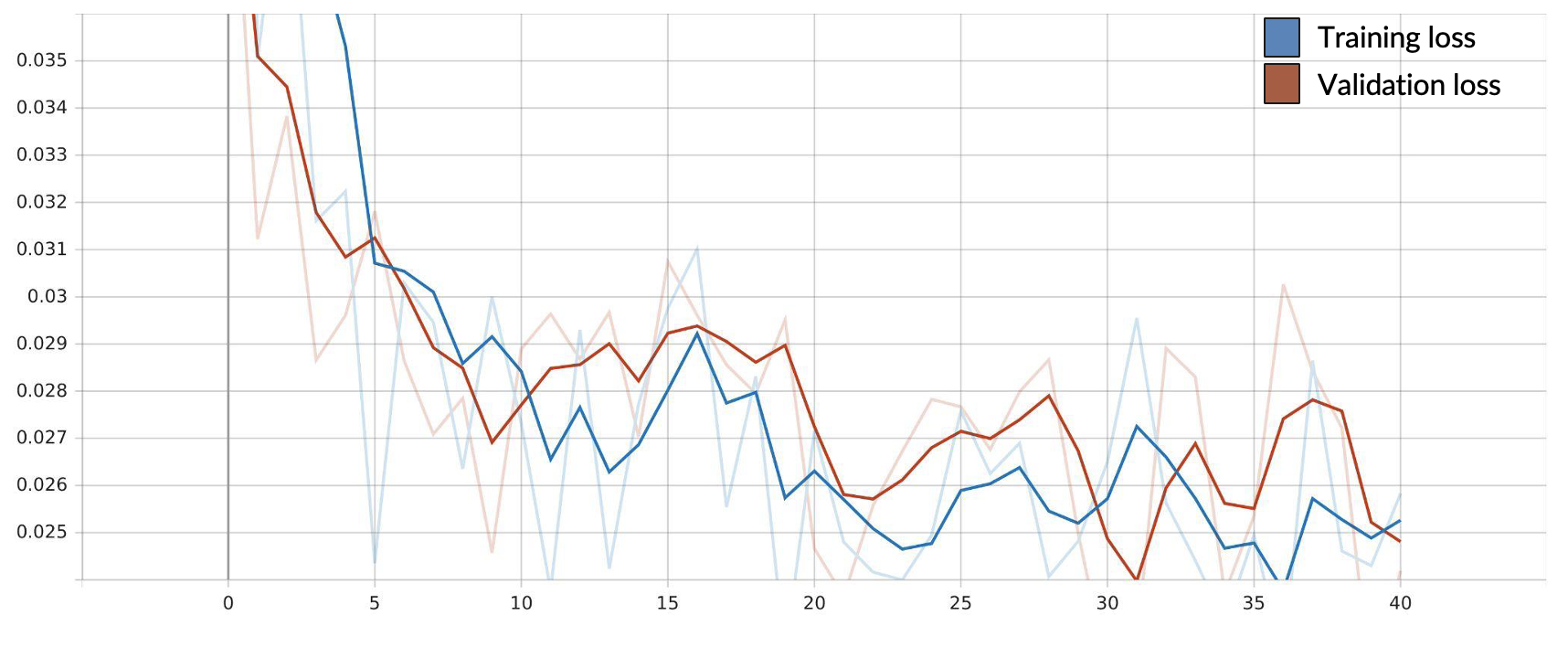}
    \caption{Plot of the training and validation losses of the raindrop detector. The plot shows that both training and validation losses decrease as the number of epochs increases. It indicates the success of model learning even though the validation loss is greater than the training loss on average}
    \label{fig:train_val_loss}
\end{figure}

For this part of the experiment, we trained our raindrop detector. Figure \ref{fig:train_val_loss} shows the plot of the training and validation losses. From this plot, we can see that the training and validation losses decrease in value as the number of epochs increases.

\begin{figure}
    \centering
    \includegraphics[width=\textwidth]{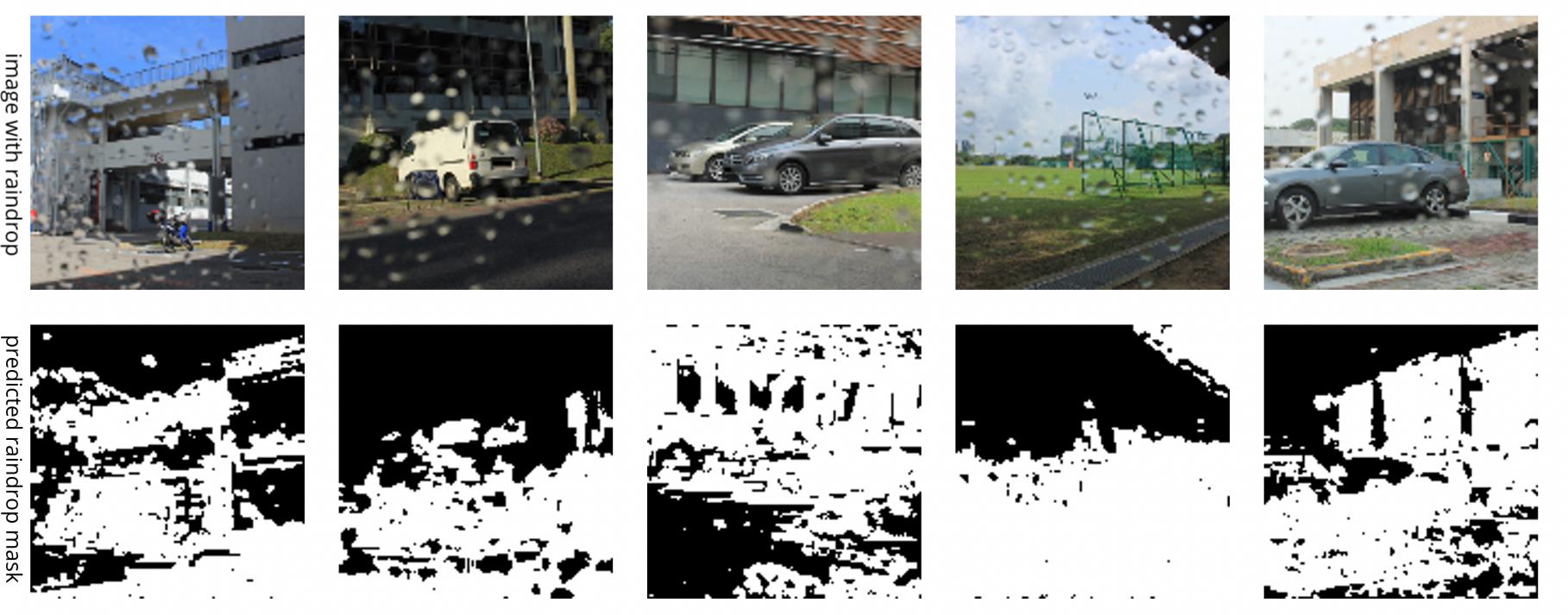}
    \caption{Samples of mask prediction by the raindrop detector applied on the Raindrop dataset. We can see visually that the masks are of poor quality. It indicates that the model overfits to the Cityscapes dataset}
    \label{fig:predicted_raindrop_mask}
\end{figure}

We then applied our raindrop detector to the Raindrop dataset. Figure \ref{fig:predicted_raindrop_mask} shows mask predictions obtained by the detector on five sample images. From this result, we can see that the mask quality is poor and does not represent the true areas containing raindrops. We analyzed the model's behavior and found that this problem was caused by domain shifting due to the change of dataset from Cityscapes to Raindrop. 

\subsubsection{Background Reconstruction}
\begin{figure}
    \centering
    \includegraphics[width=\textwidth]{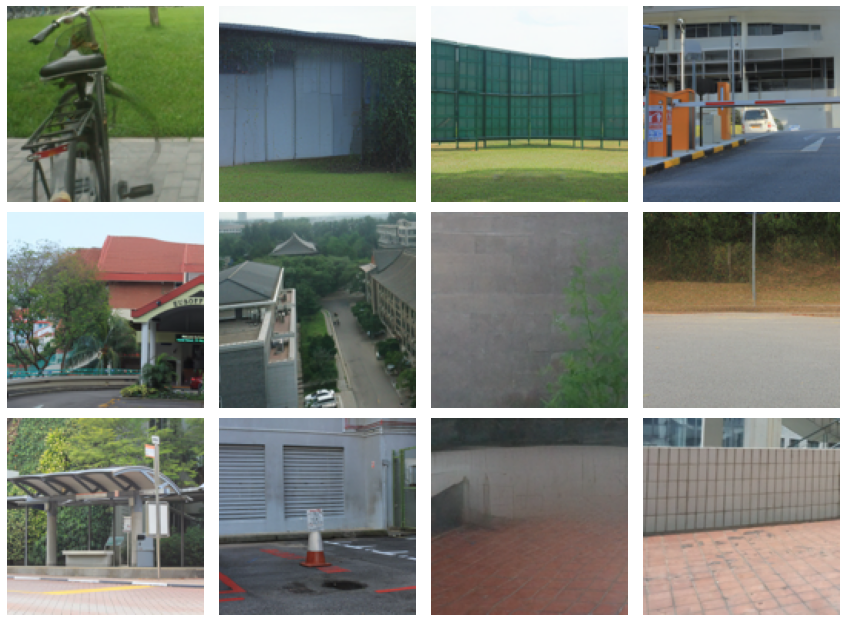}
    \caption{Novel clean images generated by a DDPM that is pre-trained on clean images of the Raindrop dataset. We can evaluate visually that these images follow the same distribution as the real clean images (cf. Figure \ref{fig:raindrop_images})}
    \label{fig:ddpm_sample_images}
\end{figure}

After we trained the DDPM model on clean images of the Raindrop dataset, we test the inference capabilities of the model. Figure \ref{fig:ddpm_sample_images} shows samples of novel images generated by the models.

\section{Conclusion and Future Work}
In this paper, we proposed a novel method to remove raindrops from a single image based on conditional diffusion models. Our method consists of raindrop detection and background reconstruction. From our experiments, we found that the residual mask generators provide more solid masks than the raindrop detection model. We argue that domain shifting (due to the change of dataset from Cityscapes to Raindrop) is the cause of this problem as it decreases model performance. Also, we are not able to perform an effective background reconstruction as a result of the lack of a proper mask. Forcing the background reconstruction to be performed on poor masks will lead to unstable reconstructions. In the future, we plan to explore more avenues of research in which we can generate better masks.


\clearpage
\bibliographystyle{unsrt}
\bibliography{bibs/references_pcl, bibs/DIffusion, bibs/RaindropRemoval, bibs/CNN, bibs/commons}

\end{document}